\documentclass[11pt]{article}
\usepackage{coling2016}
\usepackage{times}
\usepackage{url}
\usepackage{latexsym}
\usepackage{amsmath}

\usepackage{xargs}                      % Use more than one optional parameter in a new commands
\usepackage[pdftex,dvipsnames]{xcolor}  % Coloured text etc.
\usepackage{multirow}       % multirows in tables
\usepackage[colorinlistoftodos,prependcaption,textsize=tiny]{todonotes}
\newcommandx{\unsure}[2][1=]{\todo[linecolor=red,backgroundcolor=red!25,bordercolor=red,#1]{#2}}
\newcommandx{\change}[2][1=]{\todo[linecolor=blue,backgroundcolor=blue!25,bordercolor=blue,#1]{#2}}
\newcommandx{\info}[2][1=]{\todo[linecolor=OliveGreen,backgroundcolor=OliveGreen!25,bordercolor=OliveGreen,#1]{#2}}
\newcommandx{\improvement}[2][1=]{\todo[linecolor=Plum,backgroundcolor=Plum!25,bordercolor=Plum,#1]{#2}}
\newcommandx{\thiswillnotshow}[2][1=]{\todo[disable,#1]{#2}}

\title{N-gram and Neural Language Models \\ for Discriminating Similar Languages}

\author{Andre Cianflone \and Leila Kosseim \\
  Dept. of Computer Science \& Software Engineering \\
  Concordia University\\
  {\tt \{a\_cianfl|kosseim\}@encs.concordia.ca}}

\date{}

\begin{document}
\maketitle
\begin{abstract}
This paper describes our submission (named {\tt clac}) to the 2016 Discriminating Similar Languages (DSL) shared task. We participated in the closed Sub-task 1 (Set A) with two separate machine learning techniques. The first approach is a character based Convolution Neural Network with a bidirectional long short term memory (BiLSTM) layer (CLSTM), which achieved an accuracy of 78.45\% with minimal tuning. The second approach is a character-based n-gram model. This last approach achieved an accuracy of 88.45\% which is close to the accuracy of 89.38\% achieved by the best submission, and allowed us to rank \#7 overall. 
\end{abstract}

\blfootnote{This work is licensed under a Creative Commons Attribution 4.0 International Licence. Licence details:
http://creativecommons.org/licenses/by/4.0/}

%%%%%%%%%%%%%%%%%%%%%%%%%%%%%%%%%%%%%%%%%%%%%%%%%%%%%%%%%%%%%%%%%%%%%%%%%%%%%%%
% INTRODUCTION
%%%%%%%%%%%%%%%%%%%%%%%%%%%%%%%%%%%%%%%%%%%%%%%%%%%%%%%%%%%%%%%%%%%%%%%%%%%%%%%
\section{Introduction}
\label{sec:intro}

Discriminating between languages is often the first task to many natural language applications (NLP), such as  machine translation or information retrieval. Current approaches to address this problem achieve impressive results in ideal conditions: a small number of unrelated or dissimilar languages, enough training data and long enough sentences.  For example, Sim{\~o}es et al. achieved an accuracy of 97\% on the discrimination of 25 languages in TED talks~\cite{simoes2014language}. However, in the case of discriminating between similar languages  or dialects, such as French Canadian and European French, or Spanish varieties, the task is more challenging \cite{goutte-leger:2015:LT4VarDial}. This problem is addressed specifically in the DSL shared task at VarDial 2016 (DSL 2016). In comparison to results from Sim{\~o}es et al. who achieved a 97\% accuracy, the best performing system at DSL 2016 achieved only an 89.38\% accuracy. 

This paper describes our system and submission at the DSL 2016 shared task.   
The shared task is split into two main sub-tasks. Sub-task 1 aims at discriminating between similar languages and national language varieties; whereas Sub-task 2 focuses on Arabic dialect identification. 
We will only describe the specifics of Sub-task~1, for which we submitted results. For Sub-task~1, participants could chose between the closed submission, where only the use of the DSL Corpus Collection, provided by the organisers (see Section~\ref{sec:dataset}), was allowed; or the open task which permitted the use of any external data for training. Participants could also submit runs for two different data sets:  Set A, composed of newspaper articles, and Set B, composed of social media data. We only participated in the closed Sub-task~1 using Set~A.
Hence, our task was to  discriminate between 12 similar languages and national language varieties using only the newspaper articles provided in the DSL corpus as training set.
For a full description of all sub-tasks, see the overview paper~\cite{dsl2016}, which also discusses data and results for all participants.  

 It was our first participation to the DSL task, and registered late to the shared task. Hence our system is the result of a 3 person-week effort. We started with very little existing code. 
We had experimented previously with neural language models (NLM) and wanted to evaluate their applicability to this task. In addition, we believed that a convolutional plus long-short term memory network (CLSTM) would be appropriate for the task given their success in several other NLP tasks (see Section~\ref{sec:related} for details).
 In the end, we managed to submit 3 runs: {\tt run1} and {\tt run2} consist of standard character-based n-gram models; while  {\tt run3} is the CLSTM.   Our best performance was achieved by {\tt run1}, with an accuracy of 88.45\% ranking it 7\textsuperscript{th} among the 17 participants, and arriving very close to the top run which had an accuracy of 89.38\%.  Alas, our {\tt run3}, the CLSTM, attained an accuracy of 78.45\% but benefited from very minimal tuning.  

% To alleviate our text, when referring to ``shared task'', we are referring exclusively to ``closed Sub-task 1 set A'', unless otherwise specified.

%  We developed a character n-gram model as a baseline and experimented with the size of n with the training set given. However, the simpler n-gram model performed consistently better than the neural model. 

%%%%%%%%%%%%%%%%%%%%%%%%%%%%%%%%%%%%%%%%%%%%%%%%%%%%%%%%%%%%%%%%%%%%%%%%%%%%%%%
% RELATED WORK
%%%%%%%%%%%%%%%%%%%%%%%%%%%%%%%%%%%%%%%%%%%%%%%%%%%%%%%%%%%%%%%%%%%%%%%%%%%%%%%
\section{Related Work}
\label{sec:related}

Through the years, statistical language identification has  received much attention in Natural Language Processing.  The standard technique of character n-gram modeling has traditionally been very successful for this application~\cite{Cavnar94n-gram-basedtext}, but other statistical approaches such as Markov models over n-grams~\cite{Dunning94statisticalidentification}, dot products of word frequency vectors~\cite{Damashek}, and string kernels  in support vector machines~\cite{Kruengkrai05languageidentification} have also provided impressive results.  However, as noted by~\cite{Baldwin:2010:LIL:1857999.1858026}, more difficult situations where languages are similar, less training data is available or the text to identify is short can significantly degrade performance.  This is why, more recently, much effort has addressed more difficult questions such as the language identification of related languages in social media texts (e.g.~\cite{arkaitzzubiaga2014overview}) and the discrimination of similar languages (e.g.~\cite{zampieri2015overview,dsl2016}).  

The second Discriminating Similar Languages shared task (DSL 2015) aimed to discriminate between 15 similar languages and varieties, with an added ``other'' category. At this shared task, the best accuracy was 95.54\% and was achieved by \cite{malmasi-dras:2015:LT4VarDial}. The authors used two classes of features: character n-grams (with n=1 to 6) and word n-grams (with n=1 to 2). Three systems were submitted for evaluation. The first was a single Support Vector Machine (SVM) trained on the features above; while the other two systems were ensemble classifiers, combining the results of various classifiers with a mean probability combiner. A second team at DSL 2015 relied on a two-stage process, first predicting the language group and then the specific language variant \cite{goutte-leger:2015:LT4VarDial}. This team achieved an accuracy of  95.24\%. As~\cite{goutte2016discriminating} note, many other  techniques were also used for the task, such as TF-IDF and SVM, token-based backoff, prediction by partial matching with accuracies achieving between 64.04\% and 95.54\%. An interesting experiment at DSL-2015 consisted in having two versions of the corpora, where one corpus was the original newspaper articles; while the other substituted named entities with placeholders. The aim was to evaluate how strong a clue named entities are in the identification of language varieties. By relying heavily on geographic names, for example, which are highly correlated to specific nations, it was thought that accuracy would increase significantly. However, surprisingly, accuracy on the modified data set was only 1 to 2 percentage points lower than the original data set for all systems~\cite{goutte2016discriminating}. 

Given the recent success of  Recurrent Neural Networks in many NLP tasks, such as machine translation~\cite{bahdanau2014neural} and image captioning~\cite{karpathy2015deep}, we believed that an interesting approach for the DSL task would be to use solely characters as inputs, and add the ability to find long-distance relations within texts. Neural models are quite efficient at abstracting word meaning into a dense vector representation. Mikolov et al. for example, developed an efficient method to represent syntactic and semantic word relationships through a neural network~\cite{mikolov2013distributed} and the resulting vectors can be used in a variety of NLP tasks. 
For certain NLP tasks however, Convolutional Neural Networks (ConvNets), extensively studied in computer vision, have been shown to be effective for text classification. For example,~\cite{zhang2015character} experimented with ConvNets on commonly used language data sets, such as topic classification and polarity detection. A key conclusion of their study is that traditional methods, such as n-grams, work best for small data sets, whereas character ConvNets work best for data sets with millions of instances. Since the DSL data set contained a few thousand instances (see Section~\ref{sec:dataset}), we decided to give it a try.  Further, it has been shown recently that augmenting ConvNets with Recurrent Neural Networks (RNNs) is an effective way to model word sequences~\cite{kim2016character}, \cite{choi2016convolutional}. For this reason, we developed a neural model based on the latter method.

%%%%%%%%%%%%%%%%%%%%%%%%%%%%%%%%%%%%%%%%%%%%%%%%%%%%%%%%%%%%%%%%%%%%%%%%%%%%%%%
% METHODOLOGY
%%%%%%%%%%%%%%%%%%%%%%%%%%%%%%%%%%%%%%%%%%%%%%%%%%%%%%%%%%%%%%%%%%%%%%%%%%%%%%%
\section{Data Set}
\label{sec:dataset}

Because we participated in the closed task, we only used the DSL Corpus Collection (Set A)~\cite{tan:2014:BUCC} provided by the organisers.  The data set contained 12 languages organised into 5  groups: two groups of similar languages and three of national language varieties.

\begin{tabular} {ll}
\\
 Group 1:& Similar languages: Bosnian, Croatian, and Serbian\\
 Group 2:& Similar languages: Malay and Indonesian\\
 Group 3:& National varieties of Portuguese: Brazil and Portugal\\
 Group 4: &National varieties of Spanish: Argentina, Mexico, and Spain\\
Group 5: &National varieties of French: France and Canada\\
\\
\end{tabular}

Table~\ref{tab:dataset} illustrates statistics of the shared task data set.  As shown in the table, the data set is equally divided into 12 similar languages and national language varieties with 18,000 training instances for each language.  On average, each instance is 35 tokens long and contain 219 characters.

\begin{table}[h]
\center
\scalebox{0.9}{
\begin{tabular}{|c|r|llrrrcc|}
\hline
 \textbf{Group} & \multicolumn{2}{l}{\textbf{Language}}     & \textbf{Code} & \textbf{Train.}  & \textbf{Dev.}   & \textbf{Test}   & \textbf{Av. \# char.} & \textbf{Av. \# token}\\
\hline
 1              & 1& Bosnian               & bs            & 18,000           & 2,000           & 1,000           & 198                & 31\\
                & 2& Croatian              & hr            & 18,000           & 2,000           & 1,000           & 240                & 37\\
                & 3& Serbian               & sr            & 18,000           & 2,000           & 1,000           & 213                & 34\\
\hline
 2              & 4& Malaysian             & my            & 18,000           & 2,000           & 1,000           & 182                & 26\\
                & 5& Indonesian            & id            & 18,000           & 2,000           & 1,000           & 240                & 34\\
\hline
 3              & 6& Spanish (Argentina)   & es-AR         & 18,000           & 2,000           & 1,000           & 254                & 41\\
                & 7& Spanish (Spain)       & es-ES         & 18,000           & 2,000           & 1,000           & 268                & 45\\
                & 8& Spanish (Mexico)      & es-MX         & 18,000           & 2,000           & 1,000           & 182                & 31\\
\hline
 4              & 9& Portuguese (Brazil)   & pt-BR         & 18,000           & 2,000           & 1,000           & 241                & 40\\
                & 10& Portuguese (Portugal) & pt-PT         & 18,000           & 2,000           & 1,000           & 222                & 36\\
\hline
 5              & 11& French (Canada)       & fr-CA         & 18,000           & 2,000           & 1,000           & 175                & 28\\
                & 12& French (France)       & fr-FR         & 18,000           & 2,000           & 1,000           & 216                & 35\\
\hline
\multicolumn{3}{|l} {\textbf{Total}}        &               & \textbf{216,000} & \textbf{24,000} & \textbf{12,000} & \textbf{219}       & \textbf{35} \\

\hline
\end{tabular}}
\caption{Statistics of DSL 2016 Data set A. We list the number of instances across languages for the Training, Development and Test sets. The last two columns represent the average number of characters and average number of tokens of the training set.}
\label{tab:dataset}
\end{table}

Since the results of our CLSTM model (See Section~\ref{sec:clstm}) were lower than expected during the development phase, we attempted to increase the size of the training set. Using the data set from DSL-2015, we could find additional training data for most languages, with the exception of French.  We therefore attempted to use publicly vailable corpora for French. For Canadian French, we used the Canadian Hansard\footnote{http://www.isi.edu/natural-language/download/hansard/}; whereas for France French, we used the French monolingual news crawl data set (2013 version) from the ACL 2014 Ninth Workshop on Statistical Machine Translation\footnote{http://www.statmt.org/wmt14/translation-task.html}. However, upon closer investigation, this last corpus clearly contained non-French news content, heavily referencing locations and other international entities. Additionally, the majority of the Canadian French Hansard is translated from English, possibly not being representative of actual Canadian French. We experimented with these two additional data sets, but the accuracy of our models was far from our closed task equivalent. Given our short development time, we decided to drop the open task, and train our models on only the given DSL 2016 Data Set A. 

\section{Methodology}
\label{sec:method}

As indicated in Section~\ref{sec:intro}, we experimented with two main approaches: a standard n-gram model to use as baseline, and  a convolution neural network (ConvNet) with bidirectional long-short term memory recurrent neural network (BiLSTM), which we refer to as CLSTM. 

\subsection{N-gram Model}
\label{sec:ngram}

Our baseline is a standard text-book character-based n-gram model \cite{jurafsky2014speech}. Because we used a simple baseline, the same unmodified character set (including no case-folding) is used for both of our approaches, for easier later comparison. During training, the system calculates the frequency of each n-gram for each language. 
Then, at test time, the model computes a probability distribution over all possible languages and selects the most probable language as the output. Unseen n-grams were smoothed with additive smoothing with $\alpha =0.1$. As discussed in Section~\ref{sec:results}, surprisingly, this standard approach was much more accurate than our complex neural network. We experimented with different values for $n$ with the development set given (see Section~\ref{sec:dataset}).
As table~\ref{tab:ngram-validation} shows, the accuracy peaks at sizes $n=7$ and $n=8$; while larger n-grams degrade in performance and explode in memory use. The curse of dimensionality seriously limits this type of approach. 
\begin{table}[h]
\center
\begin{tabular}{|cc|}
 \hline
 \textbf{N-gram size} & \textbf{Accuracy} \\
 \hline
 1     & 0.5208   \\
 2     & 0.6733   \\
 3     & 0.7602   \\
 4     & 0.7523   \\
 5     & 0.8035   \\
 6     & 0.8303   \\
\textbf{7}      & \textbf{0.8424 }   \\
 \textbf{8}      & \textbf{0.8474 }   \\
\hline
\end{tabular}
\caption{Accuracy across n-grams of sizes 1 to 8 with the development Set A.}
\label{tab:ngram-validation}
\end{table}

\subsection{Convolution Neural Network with Long Short Term Memory (CLSTM)}
\label{sec:clstm}

Our second approach is a Convolution Neural Network with a Bidirectional Long Short Term Memory layer (CLSTM). The  goal of this approach was to build a single neural model without any feature engineering, solely taking the raw characters as input. Using characters as inputs has the added advantage of detecting language patterns even with little data available. For example, a character based neural model can predict the word {\em running} as being more likely to be in English than {\em courir} if it has seen the word {\em run} in English training texts. In a word based model that has not seen the word in this form, {\em running} would be represented as a random vector. Given the heavy computational requirements of training neural models and the limited time we had, we could not develop an ensemble neural model system, which could combine the strength of diverse models.  

The input to the model is the raw text where each character in an instance has been mapped to its one-hot representation.  Each character is therefore encoded as a vector of dimension $d$, where $d$ is a function of the maximum number of unique characters in the corpus. Luckily, the languages share heavily in alphabets and symbols, limiting $d$ to 217. A fixed number of characters $l$ is chosen from each instance. Since our texts are relatively short, as observed by the character average column in Table \ref{tab:dataset}, we set $l$ to 256. Shorter texts are zero padded, while longer instances are cut after the first 256 characters. Our input matrix $A$ is thus a $d \times l$ matrix where elements $A_{ij} \in \{0,1\}$. 
The input feeds into three sequences of convolutions and max-pooling. We used temporal max-pooling, the 1D version equivalent in computer vision.  Our ConvNet parameters are heavily based on~\cite{zhang2015character}'s empirical research  who observed that the temporal max-pooling technique is key to deep convolutional networks with text.  We further improved results on our development set by stacking the ConvNet with a Bidirectional LSTM (BiLSTM). The BiLSTM effectively takes the output of the ConvNet as its input. As shown in Table~\ref{tab:nlm}, the two LSTM layers are merged by concatenation and followed by a fully connected layer with 1024 units. ReLU is used as activation function and loss is measured on cross-entropy and optimized with the Adam algorithm \cite{kingma2015method}. The system is built as a single neural network with no pre-training. We could not test much wider networks due to lack of computing capability. However, as experienced by~\cite{zhang2015character}, it seems that much wider networks than our own would result in little, if any, performance improvement.  The model is built in Keras\footnote{https://keras.io/} and TensorFlow\footnote{https://www.tensorflow.org/}.

\begin{table}[h]
\center
\begin{tabular}{|c|lrcc|}
 \hline
\textbf{Layer}  & \textbf{Type} & \textbf{Features} & \textbf{Kernel} & \textbf{Max-pooling} \\
 \hline
1 & Convolutional & 256 & 7 & 3 \\
2 & Convolutional & 256 & 7 & 3 \\
3 & Convolutional & 256 & 3 & 3 \\
4 & LSTM (left) & 128 & - & - \\
5 & LSTM (right) & 128 & - &  -\\
6 & Dense & 1024 &-  & - \\
\hline
\end{tabular}
\caption{Layers used in our neural network. The Features column represents the number of filters for the convolutional layers and hidden units for LSTM and Dense layers. Layers 4 and 5 are merged by concatenation to form the BiLSTM layer. Dropout was added between layer 6 and the output layer (not listed in the table).}
\label{tab:nlm}
\end{table}

With the development set provided, the accuracy of the CLSTM approach reached 82\% on average which was below but comparable to the n-gram model. Additionally, our tests on the development set showed that adding the BiLSTM on top of our ConvNet does indeed increase performance. We were able to improve accuracy by 2 to 3\% on average, with little additional computing time. 

%%%%%%%%%%%%%%%%%%%%%%%%%%%%%%%%%%%%%%%%%%%%%%%%%%%%%%%%%%%%%%%%%%%%%%%%%%%%%%%
% RESULTS
%%%%%%%%%%%%%%%%%%%%%%%%%%%%%%%%%%%%%%%%%%%%%%%%%%%%%%%%%%%%%%%%%%%%%%%%%%%%%%%
\section{Results and Discussion}
\label{sec:results}
% 4.2 show table of results & how it compares to other teams

% For those languages that the n-gram did poorly, did the NLM do well!!!????
% mention the new language pair

We submitted 3 runs for the closed test Set A: 
 {\tt run1} -- the N-gram of size 7,
{\tt run2} -- the N-gram of size 8, and 
{\tt run3} -- the CLSTM model.  Table~\ref{tab:results-A-closed} shows the overall results of all 3 runs on the official test set. As the table shows, the standard n-gram model significantly outperformed the CLSTM model. It is interesting to note that the difference between the two n-grams is negligible. This was also observed during training (see Section~\ref{sec:ngram}). Recall from Table~\ref{tab:ngram-validation} that the accuracy peaked at sizes $n=7$ and $n=8$ on the development set reaching 84.74\%. The 3.71\% increase with the test set was a welcome surprise. On the other hand, the CLSTM performed about 3.55\% lower during the test than it did at training time, decreasing from an average of 82\% to 78.46\%.  Overall, as Table~\ref{tab:allteams} shows, our {\tt run1} (labeled {\tt clac}) ranked \#7 with respect to the best runs of all 17 participating teams.

\begin{table}[h]
\center
\begin{tabular}{|llrrrr|}
\hline
\bf Run & \bf Description & \bf Accuracy & \bf F1 (micro) & \bf F1 (macro) & \bf F1 (weighted) \\
\hline
Run 1 & N-gram 7 & 0.8845 & 0.8845 & 0.8813 & 0.8813 \\
Run 2 & N-gram 8 & 0.8829 & 0.8829 & 0.8812 & 0.8812 \\
Run 3 & CLSTM    & 0.7845 & 0.7845 & 0.7814 & 0.7814 \\

\hline
\end{tabular}
\caption{Results of our 3 submissions on test set A (closed training).}
\label{tab:results-A-closed}
\end{table}

\begin{table}[h]
\center
\begin{tabular}{|lllll|}
 \hline
{\bf Rank} & {\bf Team}              & {\bf Run}  & {\bf Accuracy} & {\bf F1 (weighted)} \\
\hline
1    & tubasfs           & run1 & 0.8938   & 0.8937 \\
2    & SUKI              & run1 & 0.8879   & 0.8877 \\
3    & GWU\_LT3           & run3 & 0.8870    & 0.8870 \\
4    & nrc               & run1 & 0.8859   & 0.8859 \\
5    & UPV\_UA            & run1 & 0.8833   & 0.8838 \\
6    & PITEOG            & run3 & 0.8826   & 0.8829 \\
{\bf 7}   & \textbf{clac}     & {\bf run1} & {\bf 0.8845}   & {\bf 0.8813} \\
8    & XAC               & run3 & 0.8790    & 0.8786 \\
9    & ASIREM            & run1 & 0.8779   & 0.8778 \\
10   & hltcoe            & run1 & 0.8772   & 0.8769 \\
11   & UniBucNLP         & run2 & 0.8647   & 0.8643 \\
12   & HDSL              & run1 & 0.8525   & 0.8516 \\
13   & Citius\_Ixa\_Imaxin & run2 & 0.8525   & 0.8501 \\
14   & ResIdent          & run3 & 0.8487   & 0.8466 \\
15   & eire              & run1 & 0.8376   & 0.8316 \\
16   & mitsls            & run3 & 0.8306   & 0.8299 \\
17   & Uppsala           & run2 & 0.8252   & 0.8239 \\
\hline
\end{tabular}
\caption{Results for all systems, data set A, closed track. Our system ``clac'' ranked 7\textsuperscript{th}.}
\label{tab:allteams}
\end{table}

Table~\ref{tab:ngram-test} shows the confusion matrix for our best  run, the N-gram of size~7. For comparative purposes, we have added the confusion matrix in Table~\ref{tab:clstm-test} for our third and lesser performing model, the CLSTM.  As shown in Tables \ref{tab:ngram-test} and \ref{tab:clstm-test}, for all language groups the N-gram performed significantly better than the CLSTM. However, with both models, misclassifications outside of a language group are sparse and statistically insignificant. This may indicate that a two-stage hierarchical process, as proposed by \cite{goutte-leger:2015:LT4VarDial}, is not necessary for the models we propose. 

 As shown in Tables \ref{tab:ngram-test} and \ref{tab:clstm-test}, the major difficulty for our models was the classification of the Spanish varieties in Group 3. It seems that the addition of Mexican Spanish is a significant challenge to discriminating national varieties of Spanish. At DSL 2015, \cite{goutte-leger:2015:LT4VarDial} were able to classify European Spanish and Argentine Spanish with an 89.4\% accuracy, lower than for other languages. Given the low variability among the best performing systems (see Table~\ref{tab:allteams}), and the lower performance with respect to previous iterations of the DSL shared task, this was likely a challenge for all systems at DSL 2016. 

\begin{table}[h]
\center
\scalebox{0.8}{
\begin{tabular}{|c|l|rrr|rr|rrr|rr|rr||c|}
\cline{3-14}
\multicolumn{2}{c}{}&\multicolumn{12}{|c|}{\bf Group} &\multicolumn{1}{c}{}\\
\cline{3-14}
\multicolumn{2}{c}{}&\multicolumn{3}{|c|}{\bf 1}&\multicolumn{2}{|c|}{\bf 2}&\multicolumn{3}{|c|}{\bf 3}&\multicolumn{2}{|c|}{\bf 4}&\multicolumn{2}{|c|}{\bf 5} &\multicolumn{1}{c}{}\\
\hline
 \textbf{Group} & \textbf{Code}   & \textbf{bs}   & \textbf{hr}   & \textbf{sr}   & \textbf{my}   & \textbf{id}   & \textbf{es-ar}  & \textbf{es-es}  & \textbf{es-mx}  & \textbf{pt-br}  & \textbf{pt-pt}  & \textbf{fr-ca}  & \textbf{fr-fr} & \textbf{F1}  \\
\hline
 \multirow{3}{*}{\bf 1} & \bf bs    & 674 & 182 & 142 &     &     & 1   &     & 1   &     &     &     &     & 0.75\\
                    & \bf hr    & 76  & 911 & 11  &     &     &     &     & 1   &     & 1   &     &     & 0.86\\
                    & \bf sr    & 54  & 15  & 928 &     & 1   &     &     &     &     & 1   &     & 1   & 0.89    \\
\hline
 \multirow{2}{*}{\bf 2} & \bf my    &     &     &     & 992 & 8   &     &     &     &     &     &     &     & 0.99\\
                    & \bf id    &     &     &     & 13  & 985 & 1   &     &     &     &     & 1   &     & 0.99\\
\hline
 \multirow{3}{*}{\bf 3} & \bf es-ar &     &     &     &     &     & 927 & 58  & 15  &     &     &     &     & 0.83\\
                    & \bf es-es &     &     &     &     &     & 92  & 875 & 29  &     & 2   &     & 2   & 0.81   \\
                    & \bf es-mx &     &     &     &     &     & 219 & 218 & 563 &     &     &     &     & 0.70\\
\hline
 \multirow{2}{*}{\bf 4} & \bf pt-br &     &     &     &     &     &     &     &     & 956 & 44  &     &     & 0.95\\
                    & \bf pt-pt &     &     &     &     &     &     &     &     & 54  & 946 &     &     & 0.95\\
\hline
 \multirow{2}{*}{\bf 5} & \bf fr-ca &     &     &     &     &     &     &     &     &     &     & 972 & 28  & 0.93   \\
                    & \bf fr-fr &     &     &     & 2   &     &     &     &     & 1   & 3   & 109  & 885 & 0.92  \\
\hline
\end{tabular}}
\caption{Confusion matrix for the n-gram of size 7, test Set A. We also add the  F1 score in the last column.}
\label{tab:ngram-test}
\end{table}

\begin{table}[h]
\center
\scalebox{0.8}{
\begin{tabular}{|c|l|rrr|rr|rrr|rr|rr||r|}
\cline{3-14}
\multicolumn{2}{c}{}&\multicolumn{12}{|c|}{\bf Group} &\multicolumn{1}{c}{}\\
\cline{3-14}
\multicolumn{2}{c}{}&\multicolumn{3}{|c|}{\bf 1}&\multicolumn{2}{|c|}{\bf 2}&\multicolumn{3}{|c|}{\bf 3}&\multicolumn{2}{|c|}{\bf 4}&\multicolumn{2}{|c|}{\bf 5} &\multicolumn{1}{c}{}\\
\hline\hline
 \textbf{Group} & \textbf{Code}   & \textbf{bs}   & \textbf{hr}   & \textbf{sr}   & \textbf{my}   & \textbf{id}   & \textbf{es-ar}  & \textbf{es-es}  & \textbf{es-mx}  & \textbf{pt-br}  & \textbf{pt-pt}  & \textbf{fr-ca}  & \textbf{fr-fr} & \textbf{F1}  \\
\hline
 \multirow{3}{*}{\bf 1} & \bf bs    & 697 & 172 & 129 &    & 1   & 1   &     &     &     &     &     &     & 0.67\\
                    & \bf hr    & 249 & 726 & 23  &    & 1   & 1   &     &     &     &     &     &     & 0.75\\
                    & \bf sr    & 130  & 43  & 826 &    &     &     &     &     & 1   &     &     &     & 0.83\\
\hline
 \multirow{2}{*}{\bf 2} & \bf my    &     &     &     & 909 & 91  &     &     &     &     &     &     &     & 0.94\\
                    & \bf id    &     &     &     & 23 & 975 &     & 1   &     & 1   &     &     &     & 0.94\\
\hline
 \multirow{3}{*}{\bf 3} & \bf es-ar &     &     &     &    & 2   & 816 & 87  & 93  & 2   &     &     &     & 0.71\\
                    & \bf es-es &     &     & 1   &    &     & 173 & 633 & 190  & 1   & 1   &     & 1   & 0.62    \\
                    & \bf es-mx &     &     &     &    &     & 304  & 309  & 385 & 1   &     &     & 1   & 0.46    \\
\hline
 \multirow{2}{*}{\bf 4} & \bf pt-br &     &     & 1   &    &     & 1   &     &     & 847 & 150  & 1   &     & 0.83\\
                    & \bf pt-pt &     & 1   &     &    &     & 4   &     &     & 183 & 811 &     & 1   & 0.83    \\
\hline
 \multirow{2}{*}{\bf 5} & \bf fr-ca &     &     &     &    &     &     &     &     &     &     & 972 & 28  & 0.90  \\
                    & \bf fr-fr & 1   &     &     &    &     &     & 1   &     & 1   & 1   & 178 & 818 & 0.88  \\
\hline
\end{tabular}}
\caption{Confusion matrix for the CLSTM, test Set A. We also add the  F1 score in the last column.}
\label{tab:clstm-test}
\end{table}

% is it different languages that wen wrong? --> then maybe each approach is best for a particular situation

% Talk about ensemble methods for NLM, as well as leveraging hierarchical information

%%%%%%%%%%%%%%%%%%%%%%%%%%%%%%%%%%%%%%%%%%%%%%%%%%%%%%%%%%%%%%%%%%%%%%%%%%%%%%%
% DISCUSSION AND CONCLUSION
%%%%%%%%%%%%%%%%%%%%%%%%%%%%%%%%%%%%%%%%%%%%%%%%%%%%%%%%%%%%%%%%%%%%%%%%%%%%%%%
\section{Conclusion}
Although, it still achieved an accuracy of 78.46\% with very little tuning and training set, we are disappointed in the performance of the CSLTM.  Based on the empirical study of \cite{zhang2015character}, character based ConvNets performed in line with traditional methods with data sets in the hundreds of thousands, and better with data sets in the millions. Since the shared task data set size was in between, it was not clear which approach would perform best. We believe that a deep neural network can outperform the traditional n-gram model for this task, but only once the data set size is dramatically increased and given more time to experiment on the network parameters and structure. Since only raw texts are necessary, i.e. containing no linguistic annotations, increasing the data set does not constitute a problem.

As future work, we would like to explore once again the open task. With the addition of Mexican Spanish, France French and Canadian French, discriminating similar languages continues to be a challenge. In Table \ref{tab:allteams} we see how the top 7 teams are within a 1\% spread, but all below 90\% accuracy. We believe that with a very large data set, a neural model could automatically learn key linguistic patterns to differentiate similar languages and possibly perform better than the current iteration of our CLSTM.

\subsection*{Acknowledgement}
The authors would  like to thank the anonymous reviewers for their feedback on the paper. This work was financially supported by a grant from the Natural Sciences and Engineering Research Council of Canada (NSERC).
%%%%%%%%%%%%%%%%%%%%%%%%%%%%%%%%%%%%%%%%%%%%%%%%%%%%%%%%%%%%%%%%%%%%%%%%%%%%%%%
% REFERENCES
%%%%%%%%%%%%%%%%%%%%%%%%%%%%%%%%%%%%%%%%%%%%%%%%%%%%%%%%%%%%%%%%%%%%%%%%%%%%%%%
%\clearpage
%\newpage

\bibliographystyle{acl}

\end{document}